\documentclass[10pt,journal,cspaper,compsoc]{IEEEtran}
\usepackage{graphicx}
\usepackage{amsmath,amssymb} 
\usepackage{bm}
\usepackage{booktabs} 
\usepackage{multirow}
\usepackage{epstopdf}
\usepackage{url}

\ifCLASSOPTIONcompsoc
\else
\fi
\ifCLASSINFOpdf
\else
\fi
\begin{document}
%
\title{CNN: Single-label to Multi-label}

%
%
%
%

\author{Yunchao Wei,
		Wei Xia,
		Junshi Huang,
        Bingbing Ni,
        Jian Dong,
        Yao Zhao,~\IEEEmembership{Senior Member,~IEEE}
        Shuicheng Yan,~\IEEEmembership{Senior Member,~IEEE}

\thanks{Yunchao Wei is with Department of Electrical and Computer Engineering, National University of Singapore, and also with the Institute of Information Science, Beijing Jiaotong University, e-mail: wychao1987@gmail.com.}
\thanks{Yao Zhao is with the Institute of Information Science, Beijing Jiaotong University, Beijing 100044, China.}
\thanks{Bingbing Ni is with the Advanced Digital Sciences Center, Singapore.}
\thanks{Wei Xia, Junshi Huang, Jian Dong and Shuicheng Yan are with Department of Electrical and Computer Engineering, National University of Singapore.}

}

%
%

\markboth{Journal of \LaTeX\ Class Files,~Vol.~6, No.~1, January~2014}%
{Shell \MakeLowercase{\textit{et al.}}: Bare Demo of IEEEtran.cls for Computer Society Journals}
%


\IEEEcompsoctitleabstractindextext{%
\begin{abstract}
Convolutional Neural Network (CNN) has demonstrated promising performance in single-label image classification tasks. However, how CNN best copes with multi-label images still remains an open problem, mainly due to the complex underlying object layouts and insufficient multi-label training images. In this work, we propose a flexible deep CNN infrastructure, called Hypotheses-CNN-Pooling (HCP), where an arbitrary number of  object segment hypotheses are taken as the inputs, then a shared CNN is connected with each hypothesis, and finally the CNN output results from different hypotheses are aggregated with max pooling to produce the ultimate multi-label predictions. Some unique characteristics of this flexible deep CNN infrastructure include: 1) no ground-truth bounding box information is required for training; 2) the whole HCP infrastructure is robust to possibly noisy and/or redundant hypotheses; 3) no explicit hypothesis label is required; 4) the shared CNN may be well pre-trained with a large-scale single-label image dataset, e.g. ImageNet; and 5) it may naturally output multi-label prediction results. Experimental results on Pascal VOC2007 and VOC2012 multi-label image datasets well demonstrate the superiority of the proposed HCP infrastructure over other state-of-the-arts. In particular, the mAP reaches 84.2\% by HCP only and 90.3\% after the fusion with our complementary result in~\cite{2012-Shuicheng} based on hand-crafted features on the VOC2012 dataset, which significantly outperforms the state-of-the-arts with a large margin of more than 7\%. 
\end{abstract}

}

\maketitle

\IEEEdisplaynotcompsoctitleabstractindextext

%
\IEEEpeerreviewmaketitle

\section{Introduction}
\IEEEPARstart{S}{ingle}-label image classification, which aims to assign a label from a predefined set to an image, has been extensively studied during the past few years~\cite{2007-fei-learning,2007-griffin-caltech,2009-deng-imagenet}. For image representation and classification, conventional approaches utilize carefully designed hand-crafted features, e.g., SIFT~\cite{2004-lowe-distinctive}, along with the bag-of-words coding scheme, followed by the feature pooling~\cite{2006-lazebnik-beyond,2010-wang-locality,2010-perronnin-improving} and classic classifiers, such as Support Vector Machine (SVM)~\cite{2011-chang-libsvm} and random forests~\cite{2001-randomforests}. Recently, in contrast to the hand-crafted features, \emph{learnt} image features with deep network structures have shown their great potential in various vision recognition tasks~\cite{1990-lecun,2009-jarrett,2012-krizhevsky,2013-joint-xiaogang}. Among these architectures, one of the greatest breakthroughs in image classification is the deep convolutional neural network (CNN)~\cite{2012-krizhevsky}, which has achieved the state-of-the-art performance (with 10\% gain over the previous methods based on hand-crafted features) in the large-scale single-label object recognition task, i.e., ImageNet Large Scale Visual Recognition Challenge (ILSVRC)~\cite{2009-deng-imagenet} with more than one million images from 1,000 object categories. 

Multi-label image classification is however a more general and practical problem, since the majority of real-world images are with more than one objects of different categories. Many methods~\cite{2010-perronnin-improving,2012-qiang-hierarchical,2013-dong-subcategory} have been proposed to address this more challenging problem. The success of CNN on single-label image classification also sheds some light on the multi-label image classification problem. However, the CNN model cannot be trivially extended to cope with the multi-label image classification problem in an interpretable manner, mainly due to the following reasons. Firstly, the implicit assumption that foreground objects are roughly aligned, which is usually true for single-label images, does not always hold for multi-label images. Such alignment facilitates the design of the convolution and pooling infrastructure of CNN for single-label image classification.  However, for a typical multi-label image, different categories of objects are located at various positions with different scales and poses. For example, as shown in Figure~\ref{fig:single-multi}, for single-label images, the foreground objects are roughly aligned, while for multi-label images, even with the same label, i.e., \emph{horse and person}, the spatial arrangements of the \emph{horse} and \emph{person} instances vary largely among different images. Secondly, the interaction between different objects in multi-label images, like partial visibility and occlusion, also poses a great challenge. Therefore, directly applying the original CNN structure for multi-label image classification is not feasible.
Thirdly, due to the tremendous parameters to be learned for CNN, a large number of training images are required for the model training. Furthermore, from single-label to multi-label (with $n$ category labels) image classification, the label space has been expanded from ${n}$ to ${2}^{n}$, thus more training data is required to cover the whole label space. For single-label images, it is practically easy to collect and annotate the images. However, the burden of collection and annotation for a large scale multi-label image dataset is generally extremely high.

To address these issues and take full advantage of CNN for multi-label image classification, in this paper, we propose a flexible deep CNN structure, called Hypotheses-CNN-Pooling (HCP). HCP takes an arbitrary number of object segment hypotheses as the inputs, which may be generated by the sate-of-the-art objectiveness detection techniques, e.g., binarized normed gradients (BING)~\cite{2014-chengmm}, and then a shared CNN is connected with each hypothesis. Finally the CNN output results from different hypotheses are aggregated by max pooling to give the ultimate multi-label predictions. Particularly, the proposed HCP infrastructure possesses the following characteristics:
\begin{figure}[t]
\centering
\includegraphics[scale=0.37]{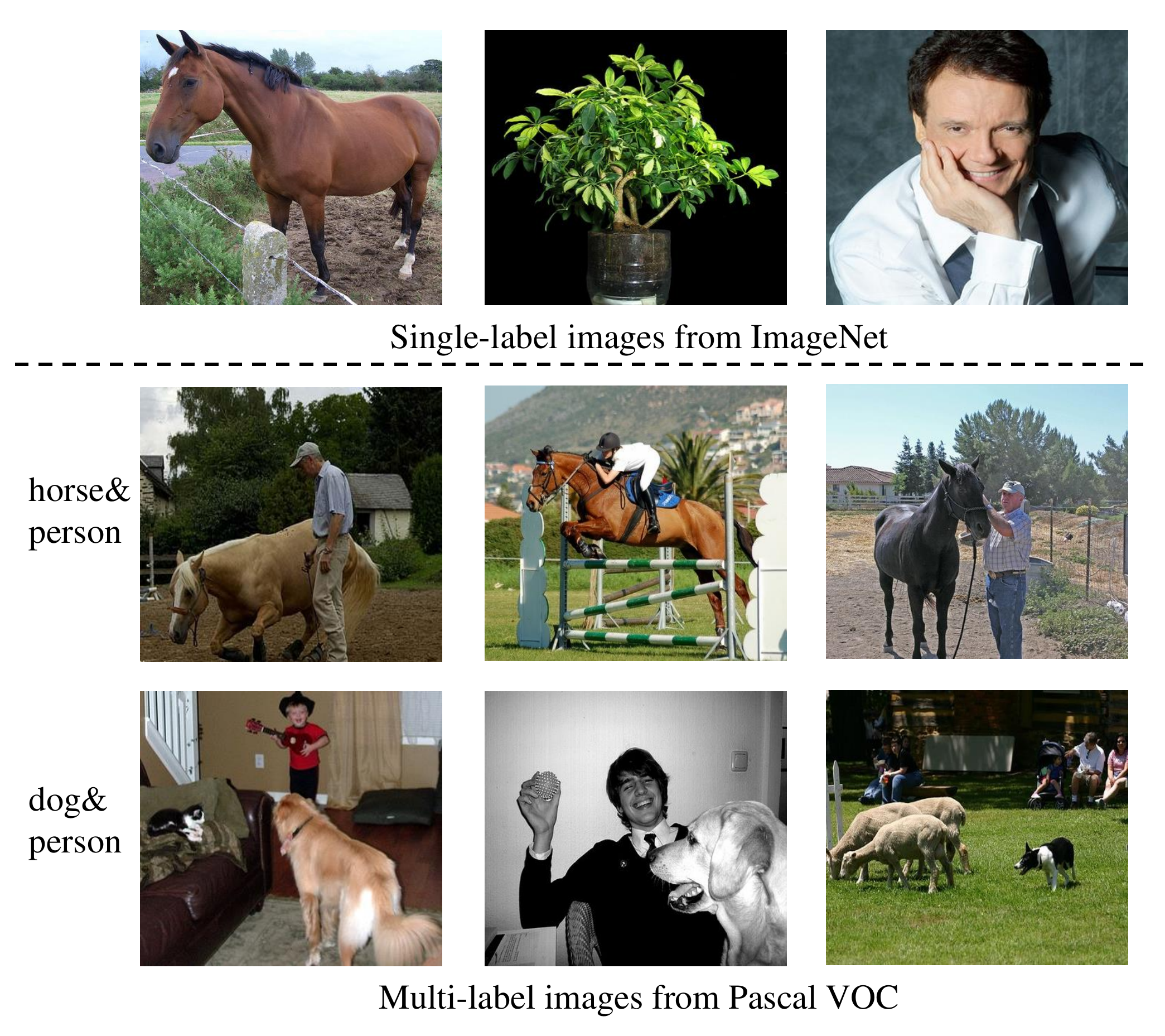}
\caption{Some examples from ImageNet~\cite{2009-deng-imagenet} and Pascal VOC 2007~\cite{2010-pascal}. The foreground objects in single-label images are usually roughly aligned. However, the assumption of object alighment is not valid for multi-label images. Also note the partial visibility and occlusion between objects in the multi-label images.}
\label{fig:single-multi}
\end{figure}
\begin{itemize}
\item[$\bullet$] \textit{No ground-truth bounding box information is required for training on the multi-label image dataset.} Different from previous works~\cite{2013-dong-subcategory,2014-chen,2013-richfeatures,2013-transferCNN}, which employ ground-truth bounding box information for training, the proposed HCP requires no bounding box annotation. Since bounding box annotation is much more costly than labelling, the annotation burden is significantly reduced. Therefore, the proposed HCP has a better generalization ability when transferred to new multi-label image datasets.\\
\item[$\bullet$] \textit{The proposed HCP infrastructure is robust to the noisy and/or redundant hypotheses.} To suppress the possibly noisy hypotheses, a cross-hypothesis max-pooling operation is carried out to fuse the outputs from the shared CNN into an integrative prediction. With max pooling, the high predictive scores from those hypotheses containing objects are reserved and the noisy ones are ignored. Therefore, as long as one hypothesis contains the object of interest, the noise can be suppressed after the cross-hypothesis pooling. Redundant hypotheses can also be well addressed by max pooling.\\
\item[$\bullet$] \textit{No explicit hypothesis label is required for training.} The state-of-the-art CNN models~\cite{2013-richfeatures,2013-transferCNN} utilize the hypothesis label for training. They first compute the Intersection-over-Union (IoU) overlap between hypotheses and ground-truth bounding boxes, and then assign the hypothesis with the label of the ground-truth bounding box if their overlap is above a threshold. In contrast, the proposed HCP takes an arbitrary number of hypotheses as the inputs without any explicit hypothesis labels. \\
\item[$\bullet$] \textit{The shared CNN can be well pre-trained with a large-scale single-label image dataset.} To address the problem of insufficient multi-label training images, based on the Hypotheses-CNN-Pooling architecture, the shared CNN can be first well pre-trained on some large-scale single-label dataset, e.g., ImageNet, and then fine-tuned on the target multi-label dataset. \\

\item[$\bullet$] \textit{The HCP outputs are intrinsically multi-label prediction results.} HCP produces a normalized probability distribution over the labels after the softmax layer, and the the predicted probability values are intrinsically the final classification confidence values for the corresponding categories.\\
\end{itemize}

Extensive experiments on two challenging multi-label image datasets, Pascal VOC 2007 and VOC 2012,  well demonstrate the superiority of the proposed HCP infrastructure over other state-of-the-arts. The rest of the paper is organized as follows. We briefly review the related work of multi-label classification in Section~\ref{sec:relatedworks}. Section~\ref{sec:hcp} presents the details of the HCP for image classification. Finally the experimental results and conclusions are provided in Section~\ref{sec:experiments} and Section~\ref{sec:conclusions}, respectively.

\begin{figure*}[t]
\centering
\includegraphics[scale=0.6]{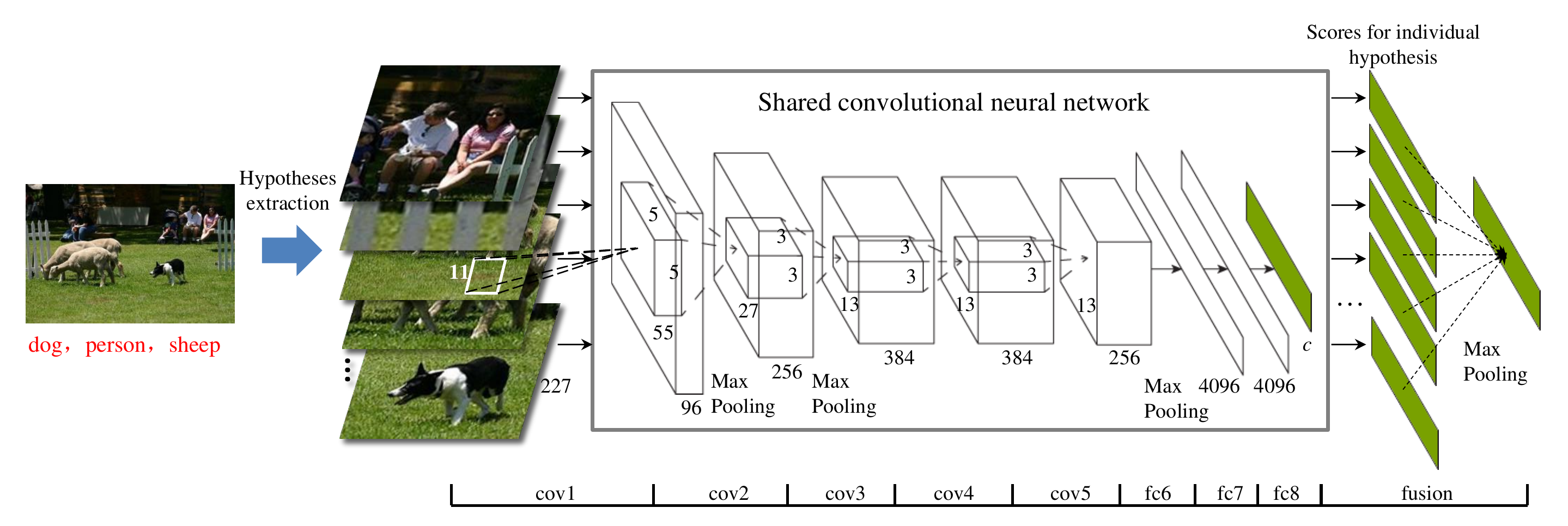}
\caption{An illustration of the infrastructure of the proposed HCP. For a given multi-label image, a set of input hypotheses to the shared CNN is selected based on the proposals generated by the state-of-the-art objectness detection techniques, e.g., BING~\cite{2014-chengmm}. The shared CNN has a similar network structure to~\cite{2012-krizhevsky} except for the layer fc8, where $c$ is the category number of the target multi-label dataset. We feed the selected hypotheses into the shared CNN and fuse the outputs into a $c$-dimensional prediction vector with cross-hypothesis max-pooling operation. The shared CNN is firstly pre-trained on the single-label image dataset, e.g., ImageNet and then fine-tuned with the multi-label images based on the squared loss function. Finally, we retrain the whole HCP to further fine-tune the parameters for multi-label image classification.}
\label{fig:framework}
\end{figure*}
\section{Related Work}
\label{sec:relatedworks}
During the past few years, many works on various multi-label image classification models have been conducted. These models are generally based on two types of frameworks: bag-of-words (BoW)~\cite{2009-harzallah,2010-perronnin-improving,2012-qiang-hierarchical,2013-dong-subcategory,2014-chen} and  deep learning~\cite{2013-transferCNN,2014-yunchao,2014-cnnfeatures}.
\subsection{Bag-of-Words Based Models}
A traditional BoW model is composed of multiple modules, e.g., feature representation, classification and context modelling. For feature representation, the main components include hand-crafted feature extraction, feature coding and feature pooling, which generate global representations for images. Specifically, hand-crafted features, such as SIFT~\cite{2004-lowe-distinctive}, Histogram of Oriented Gradients~\cite{2005-hog} and Local Binary Patterns~\cite{1996-LBP} are firstly extracted on dense grids or sparse interest points and then quantized by different coding schemes, e.g., Vector Quantization~\cite{1988-vq}, Sparse Coding~\cite{2010-wright} and Gaussian Mixture Models~\cite{2000-gmm}. These encoded features are finally pooled by feature aggregation methods, such as Spatial Pyramid Matching (SPM)~\cite{2006-lazebnik-beyond}, to form the image-level representation. For classification, conventional models, such as SVM~\cite{2011-chang-libsvm} and random forests~\cite{2001-randomforests}, are utilized. Beyond conventional modelling methods, many recent works~\cite{2009-harzallah,2011-song,2012-feifei,2012-qiang-hierarchical,2014-chen} have demonstrated that the usage of context information, e.g., spatial location of object and background scene from the global view, can considerably improve the performance of multi-label classification and object detection.

Although these works have made great progress in visual recognition tasks, the involved hand-crafted features are not always optimal for particular tasks. Recently, in contrast to hand-crafted features, \emph{learnt} features with deep learning structures have shown great potential for various vision recognition tasks, which will be introduced in the following subsection. 


\subsection{Deep Learning Based Models}
Deep learning tries to model the high-level abstractions of visual data by using architectures composed of multiple non-linear transformations. Specifically, deep convolutional neural network (CNN)~\cite{1990-lecun} has demonstrated an extraordinary ability for image classification~\cite{2009-jarrett,2004-lecun,2009-lee,2012-krizhevsky,2014-linmin} on single-label datasets such as CIFAR-10/100~\cite{2009-cifar} and ImageNet~\cite{2009-deng-imagenet}.

More recently, CNN architectures have been adopted to address multi-label problems. Gong \emph{et al.}~\cite{2014-yunchao} studied and compared several multi-label loss functions for the multi-label annotation problem based on a similar network structure to~\cite{2012-krizhevsky}. However, due to the large number of parameters to be learned for CNN, an effective model requires lots of training samples. Therefore, training a task-specific convolutional neural network is not applicable on datasets with limited numbers of training samples. 

Fortunately, some recent works~\cite{2013-jia,2013-richfeatures,2013-transferCNN,2013-overfeat,2014-cnnfeatures,2014-yunchao-vlad} have demonstrated that CNN models pre-trained on large datasets with data diversity, e.g., ImageNet, can be transferred to extract CNN features for other image datasets without enough training data. Pierre \emph{et al.}~\cite{2013-overfeat} and Razavian \emph{et al.}~\cite{2014-cnnfeatures} proposed a CNN feature-SVM pipeline for multi-label classification. Specifically, global images from a multi-label dataset are directly fed into the CNN, which is pre-trained on ImageNet, to get CNN activations as the off-the-shelf features for classification. However, different from the single-label image, objects in a typical multi-label image are generally less-aligned, and also often with partial visibility and occlusion as shown in Figure~\ref{fig:single-multi}. Therefore, global CNN features are not optimal to multi-label problems. Recently, Oquab \emph{et al.}~\cite{2013-transferCNN} and Girshick \emph{et al.}~\cite{2013-richfeatures} presented two proposal-based methods for multi-label classification and detection. Although considerable improvements have been made by these two approaches, these methods highly depend on the ground-truth bounding boxes, which may limit their generalization ability when transferred to a new multi-label dataset without any bounding box information. 

In contrast, the proposed HCP infrastructure in this paper requires no ground-truth bounding box information for training and is robust to the possibly noisy and/or redundant hypotheses. Different from~\cite{2013-transferCNN,2013-richfeatures}, no explicit hypothesis label is required during the training process. Besides, we propose a hypothesis selection method to select a small number of high-quality hypotheses (10 for each image) for training, which is much less than the number used in~\cite{2013-richfeatures} (128 for each image), thus the training process is significantly sped up.

\section{Hypotheses-CNN-Pooling}
Figure 2 shows the architecture of the proposed Hypotheses-CNN-Pooling (HCP) deep network. We apply the state-of-the-art objectness detection technique, i.e., BING~\cite{2014-chengmm}, to produce a set of candidate object windows. A much smaller number of candidate windows are then selected as hypotheses by the proposed hypotheses extraction method. The selected hypotheses are fed into a shared convolutional neural network (CNN). The confidence vectors from the input hypotheses are combined through a fusion layer with max pooling operation, to generate the ultimate multi-label predictions. In specific, the shared CNN is first pre-trained on a large-scale single-label image dataset, i.e., ImageNet and then fine-tuned on the target multi-label dataset, e.g., Pascal VOC, by using the entire image as the input. After that, we retrain the proposed HCP with a squared loss function for the final prediction.
\label{sec:hcp}
\subsection{Hypotheses Extraction}
\label{subsec:hypotheses extraction}
HCP takes an arbitrary number of object segment hypotheses as the inputs to the shared CNN and fuses the prediction of each hypothesis with the max pooling operation to get the ultimate multi-label predictions. Therefore, the performance of the proposed HCP largely depends on the quality of the extracted hypotheses. Nevertheless, designing an effective hypotheses extraction approach is challenging, which should satisfy the following criteria:

\noindent\textbf{High object detection recall rate:} The proposed HCP is based on the assumption that the input hypotheses can cover all single objects of the given multi-label image, which requires a high detection recall rate. 

\noindent\textbf{Small number of hypotheses:} Since all hypotheses of a given multi-label image need to be fed into the shared CNN simultaneously, more hypotheses cost more computational time and need more powerful hardware (e.g., RAM and GPU). Thus a small hypothesis number is required for an effective hypotheses extraction approach.

\noindent\textbf{High computational efficiency:} As the first step of the proposed HCP, the efficiency of hypotheses extraction will significantly influence the performance of the whole framework. With high computational efficiency, HCP can be easily integrated into real-time applications. 

In summary, a good hypothesis generating algorithm should generate as few hypotheses as possible in an efficient way and meanwhile achieve as high recall rate as possible. 

During the past few years, many methods~\cite{2011-liu,2013-Cheng,2013-Jiaya,2012-alexe,2012-cpmc,2013-selective} have been proposed to tackle the hypotheses detection problem. \cite{2011-liu,2013-Cheng,2013-Jiaya} are based on salient object detection, which try to detect the most attention-grabbing (salient) object in a given image. However, these methods are not applicable to HCP, since saliency based methods are usually applied to a single-label scheme while HCP is a multi-label scheme. \cite{2012-alexe,2012-cpmc,2013-selective} are based on objectness proposal (hypothesis), which generate a set of hypotheses to cover all independent objects in a given image. Due to the large number of proposals, such methods are usually quite time-consuming, which will affect the real-time performance of HCP. 

\begin{figure}[t]
\centering
\includegraphics[scale=0.29]{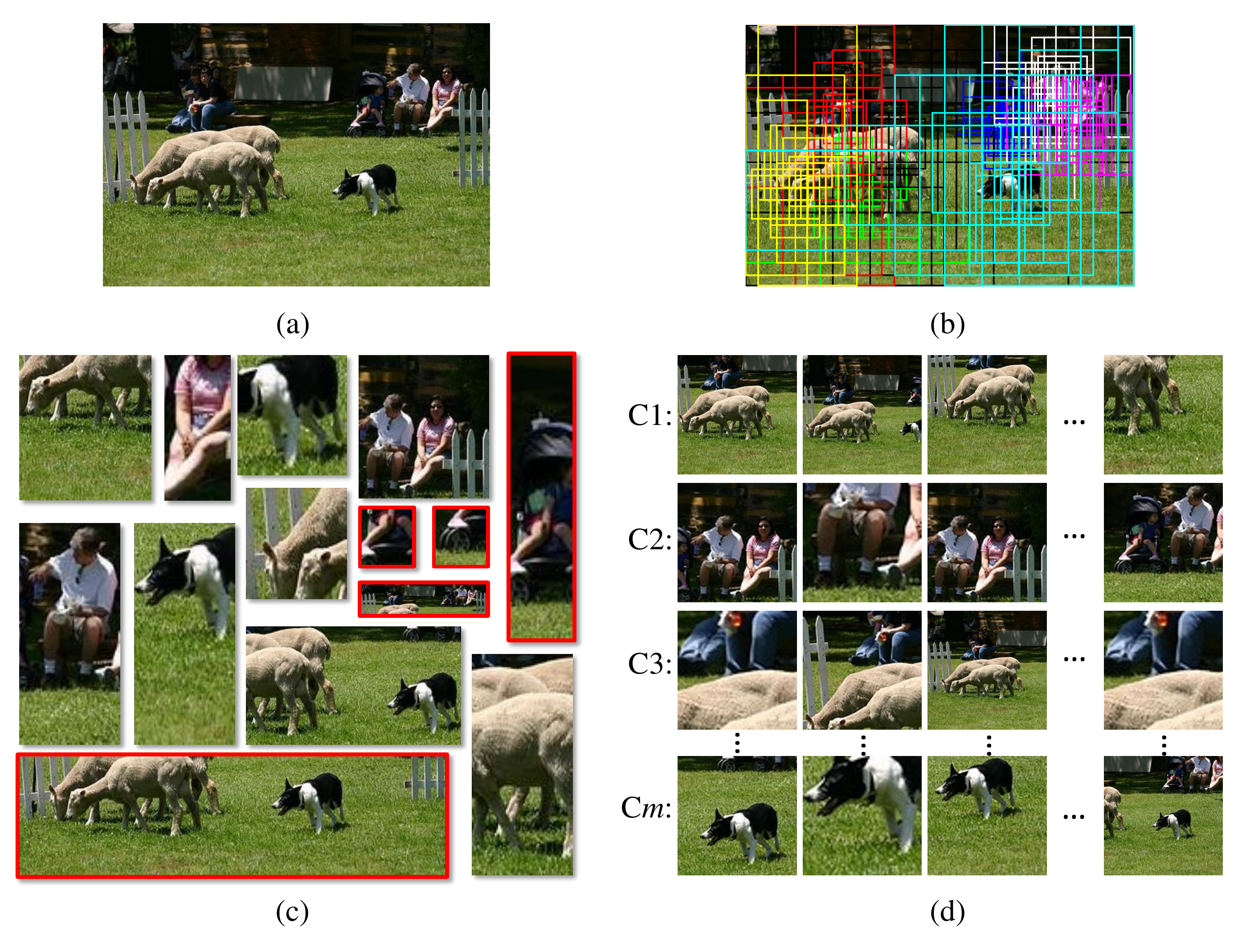}
\caption{(a) Source image. (b) Hypothesis bounding boxes generated by BING. Different colors indicate different clusters, which are produced by normalized cut. (c) Hypotheses directly generated by the bounding boxes. (d) Hypotheses generated by the proposed HS method. }
\label{fig:bing}
\end{figure}

Most recently, Cheng \emph{et al.} \cite{2014-chengmm} proposed a surprisingly simple and powerful feature called binarized normed gradients (BING) to find object candidates by using objectness scores. This method is faster (300fps on a single laptop CPU) than most popular alternatives~\cite{2012-alexe,2012-cpmc,2013-selective} and has a high object detection recall rate (96.2\% with 1,000 hypotheses). 
Although the number of hypotheses (i.e., 1,000) is very small compared with a common sliding window paradigm, it is still very large for HCP. 

To address this problem, we propose a hypotheses selection (HS) method to select hypotheses from the proposals extracted by BING. A set of hypothesis bounding boxes are produced by BING for a given image, denoted by ${H} = \{ {h_{1}},{h_{2}},...,{h_{n}}\}$, where $n$ is the hypothesis number. An $n \times n$ affinity matrix $W$ is constructed, where $W_{ij}$ ($i, j <= n$) is the IoU scores between $b_{i}$ and $b_{j}$, which can be defined as
\begin{equation}
\label{eq:W}
{W_{ij}} = \frac{{\left| {{h_i} \cap {h_j}} \right|}}{{\left| {{h_i} \cup {h_j}} \right|}},
\end{equation}
where $\left|  \cdot  \right|$ is used to measure the number of pixels. The normalized cut algorithm~\cite{2000-normalized} is then adopted to group the hypothesis bounding boxes into $m$ clusters. As shown in Figure~\ref{fig:bing}(b), different colors indicate different clusters. We empirically filter out those hypotheses with small areas or with high height/width (or width/height) ratios, as those shown in Figure~\ref{fig:bing}(c) with red bounding boxes. For each cluster, we pick out the top $k$ hypotheses with higher predictive scores generated by BING and resize them into square shapes. As a result, $mk$ hypotheses, which are much fewer than those directly generated by BING, will be selected as the inputs of HCP for each image.

\subsection{Initialization of HCP}
\label{subsection: initializeHCP}
In the proposed HCP, the architecture of the shared CNN is similar to the network described in~\cite{2012-krizhevsky}. The shared CNN contains five convolutional layers and three fully-connected layers with 60 million parameters. Therefore, without enough training images, it is very difficult to obtain an effective HCP model for multi-label classification. However, to collect and annotate a large-scale multi-label dataset is generally unaffordable. Fortunately, a large-scale single-label image dataset, i.e., ImageNet, can be used to pre-train the shared CNN for parameter initialization, since each image of multi-label is firstly cropped into many hypotheses and each hypothesis is assumed to contain at most one object based on the architecture of HCP.

However, directly using the parameters pre-trained by ImageNet to initialize the shared CNN is not appropriate, due to the following reasons. Firstly, both the data amount and the object categories between ImageNet and the target muliti-label dataset are usually different. Secondly, there exist very diverse and complicated interactions among the objects in a multi-label image, which makes multi-label classification more challenging than single-label classification. To better initialize the shared CNN, based on the pre-trained parameters by ImageNet, fine-tuning is enforced to adjust the parameters.

\begin{figure}[t]
\centering
\includegraphics[scale=0.42]{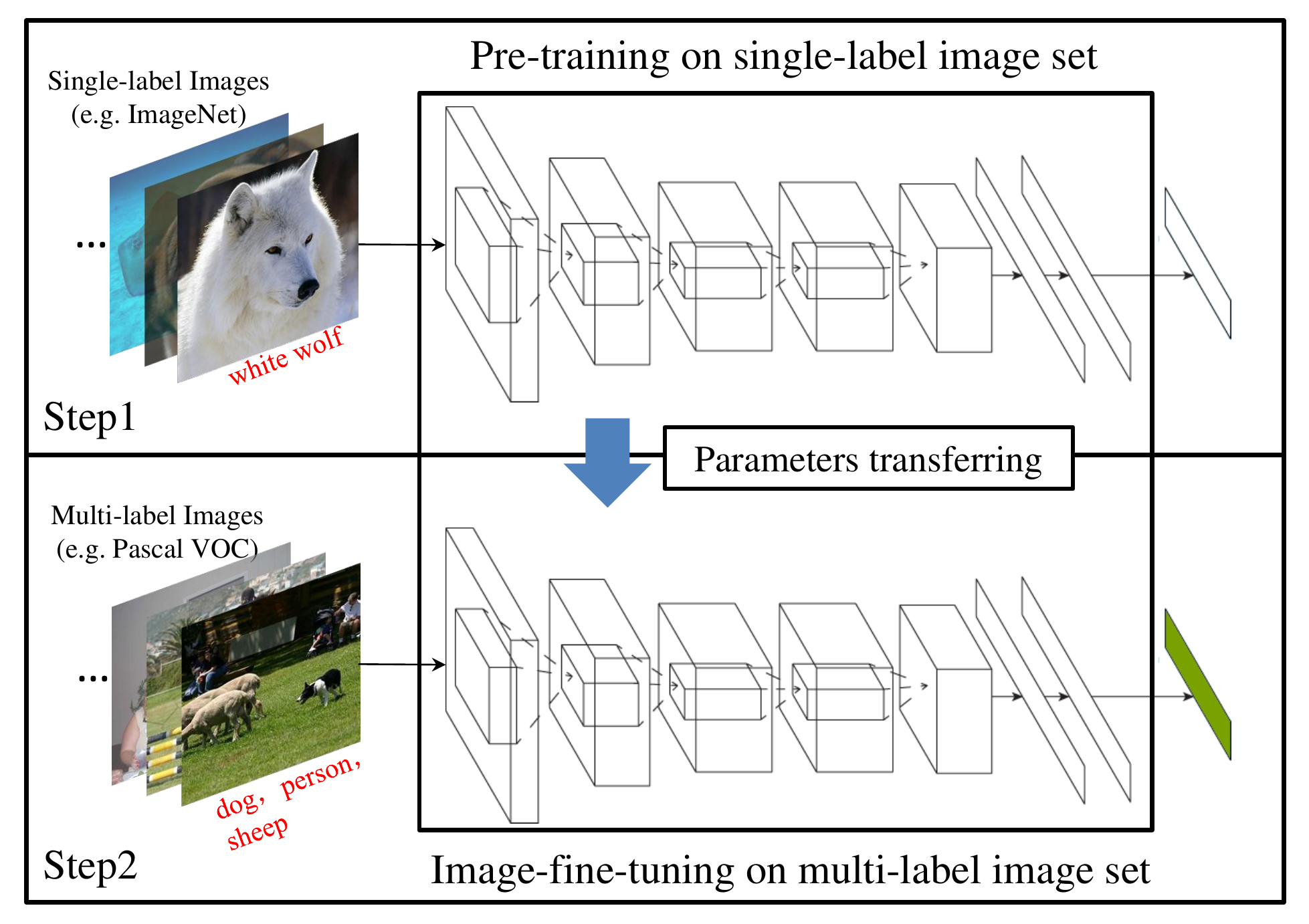}
\caption{The initialization of HCP is divided into two steps. The shared CNN is first pre-trained on a single-label image set, e.g., ImageNet and then fine-tuned on the target multi-label image set using the entire image as input. Parameters pre-trained on ImageNet are directly transferred for fine-tuning except for the last fully-connected layer, since the category numbers between these two datasets are different.}
\label{fig:finetune}
\end{figure}

As shown in Figure~\ref{fig:finetune}, the initialization process of HCP is divided into two steps. 

\noindent\textbf{Step1: Pre-training on single-label image set.} We use the ImageNet~\cite{2009-deng-imagenet} to pre-train the shared CNN. Given an image, we first resize it into 256$\times$256 pixels. Then, we extract random 227$\times$227 patches (and their horizontal reflections) from the given image and train our network based on these extracted patches. Each extracted patch is pre-processed by subtracting the image mean, and fed into the first convolutional layer of the CNN. Indicated by~\cite{2012-krizhevsky}, the output of the last fully-connected layer is fed into a 1,000-way softmax layer with the multinomial logistic regression as the loss function, to produce a probability distribution over the 1,000 classes. For all layers, we use the rectified linear units (ReLU) as the nonlinear activation function. We train the network by using stochastic gradient descent with a momentum of 0.9 and weight decay of 0.0005. To overcome overfitting, each of the first two fully-connected layers is followed by a drop-out operation with a drop-out ratio of 0.5. The learning rate is initialized as 0.01 for all layers and reduced to one tenth of the current rate after every 20 epoches (90 epoches in all).

\noindent\textbf{Step2: Image-fine-tuning on multi-label image set.} To adapt the pre-trained model on ImageNet to HCP, the entire images from a multi-label image set, e.g., Pascal VOC, are then utilized to further adjust the parameters. The image-fine-tuning (I-FT) process is similar with the pre-training except for several details listed as follows.

Each image is resized into 256$\times$256 pixels without cropping. Since the category number of Pascal VOC is not equal to that of ImageNet, the output of the last fully-connected layer is fed into a $c$-way softmax which produces a probability distribution over the $c$ class labels. Different from the pre-training, squared loss is used during I-FT. Suppose there are $N$ images in the multi-label image set, and $\bm{y}_i = [{y_{i1}},{y_{i2}},\cdots,{y_{ic}}]$ is the label vector of the  $i^{th}$ image. $y_{ij}=1$ $(j=1,\cdots,c)$ if the image is annotated with class $j$, and otherwise $y_{ij}=0$. The ground-truth probability vector of the $i^{th}$ image is defined as $\bm{{\hat p}}_i=\bm{y}_i/||\bm{y}_i||_1$ and the predictive probability vector is $\bm{{p}_i}=[{p_{i1}},{p_{i2}},\cdots,{p_{ic}}]$. And then the cost function to be minimized is defined as 
\begin{equation}
\label{eq: cost function}
J = \frac{1}{N}\sum\limits_{i = 1}^N {\sum\limits_{k = 1}^c {{{({p_{ik}} - {{\hat p}_{ik}})}^2}} } .
\end{equation}
During the I-FT process, as shown in Figure~\ref{fig:finetune}, the parameters of the first seven layers are initialized by the parameters pre-trained on ImageNet and the parameters of the last fully-connected layer are randomly initialized with a Gaussian distribution $G(\mu,\sigma)$($\mu=0, \sigma=0.01$). The learning rates of the convolutional layers, the first two fully-connected layers and the last fully-connected layer are initialized as 0.001, 0.002 and 0.01 at the beginning, respectively. We executed 60 epoches in total and decreased the learning rate to one tenth of the current rate of each layer after 20 epoches (momentum$=$0.9, weight decay$=$0.0005). 

By setting the different learning rates for different layers, the updating rates for the parameters from different layers also vary. The first few convolutional layers mainly extract some low-level invariant representations, thus the parameters are quite consistent from the pre-trained dataset to the target dataset, which is achieved by a very low learning rate (i.e., 0.001). Nevertheless, in the final layers of the network, especially the last fully-connected layer, which are specifically adapted to the new target dataset, a much higher learning rate is required to guarantee a fast convergence to the new optimum. Therefore, the parameters can better adapt to the new dataset without clobbering the pre-trained initialization. It should be noted that the I-FT is a critical step of HCP. We tried without this step and found that the performance on VOC 2007 dropped dramatically.

\subsection{Hypotheses-fine-tuning}
All the $l=mk$ hypotheses are fed into the shared CNN, which has been initialized as elaborated in Section~\ref{subsection: initializeHCP}. For each hypothesis, a $c$-dimensional vector can be computed as the output of the shared CNN. Indeed, the proposed HCP is based on the assumption that each hypothesis contains at most one object and all the possible objects are covered by some subset of the extracted hypotheses. Therefore, the number of hypotheses should be large enough to cover all possible diversified objects. However, with more hypotheses, noise (hypotheses covering no object) will inevitably increase. 

To suppress the possibly noisy hypotheses, a cross-hypothesis max-pooling is carried out to fuse the outputs into one integrative prediction. Suppose ${\bm{v}_i} (i = 1,...,l)$ is the output vector of the  $i^{th}$ hypothesis from the shared CNN and $\bm{v}_i^{(j)} (j = 1,\dots,c)$ is the $j^{th}$ component of $\bm{v}_i$. The cross-hypothesis max-pooling in the fusion layer can be formulated as
\begin{equation}
{\bm{v}^{(j)}} = \max (\bm{v}_1^{(j)},\bm{v}_2^{(j)},\dots,\bm{v}_l^{(j)}),
\end{equation}
where $\bm{v}^{(j)}$ can be considered as the predicted value for the $j^{th}$ category of the given image.

The cross-hypothesis max-pooling is a crucial step for the whole HCP framework to be robust to the noise. If one hypothesis contains an object, the output vector will have a high response (i.e., large value) on the $j^{th}$ component, meaning a high confidence for the corresponding $j^{th}$ category. With cross-hypothesis max-pooling, large predicted values corresponding to objects of interest will be reserved, while the values from the noisy hypotheses will be ignored. 

During the hypotheses-fine-tuning (H-FT) process, the output of the fusion layer is fed into a $c$-way softmax layer with the squared loss as the cost function, which is defined as Eq.~(\ref{eq: cost function}). 
Similar as I-FT, we also adopt a discriminating learning rate scheme for different layers. Specifically, we execute 60 epoches in total and empirically set the learning rates of the convolutional layers, the first two fully-connected layers and the last fully-connected layer as 0.0001, 0.0002, 0.001 at the beginning, respectively. We decrease the learning rates to one tenth of the current ones after every 20 epoches. The momentum and the weight decay are set as 0.9 and 0.0005, which are the same as in the I-FT step. 
\subsection{Multi-label Classification for Test Image}
\begin{figure}[t]
\centering
\includegraphics[scale=0.52]{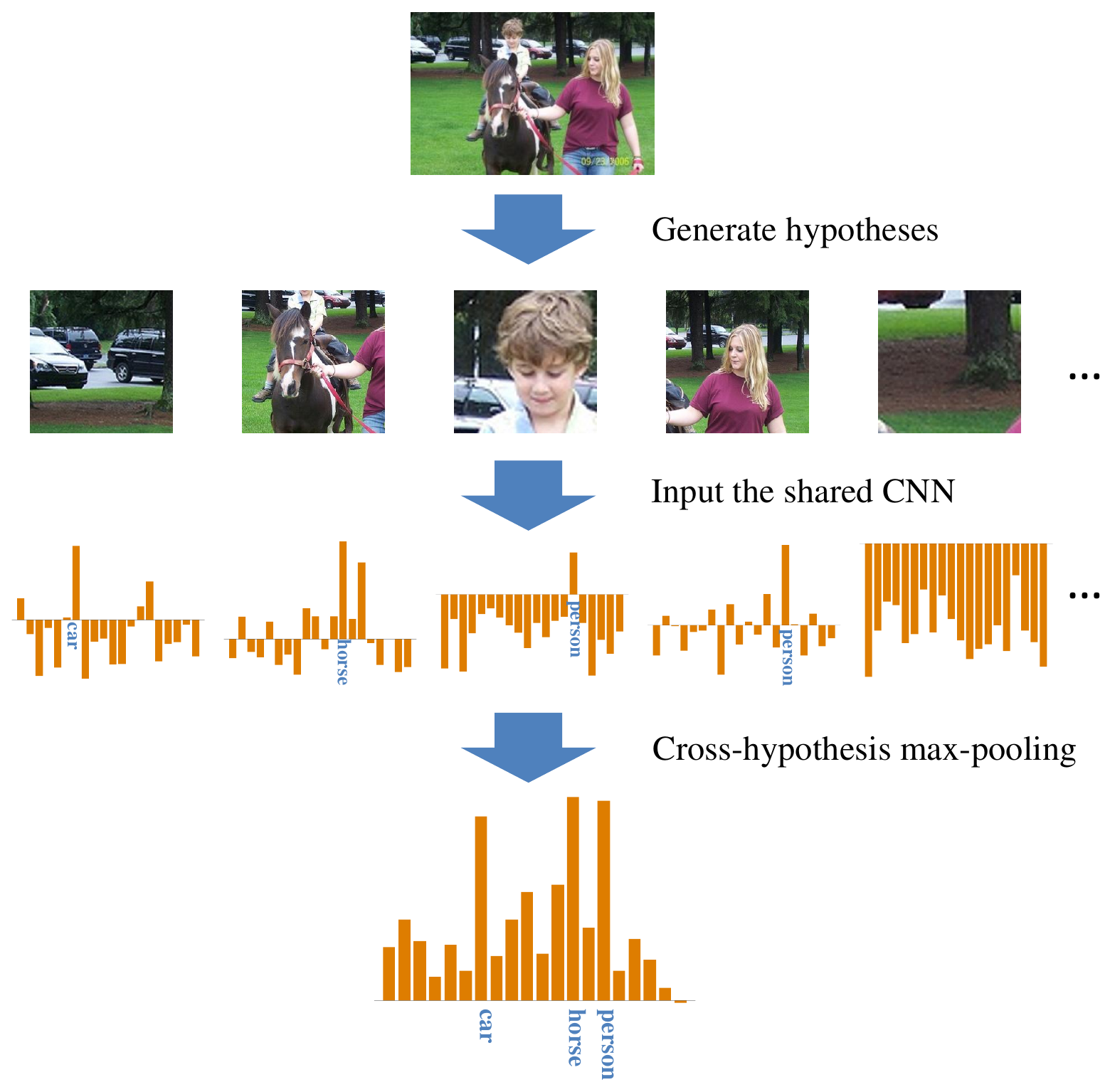}
\caption{An illustration of the proposed HCP for a VOC 2007 test image. The second row indicates the generated hypotheses. The third row indicate the predicted results for the input hypotheses. The last row is predicted result for the test image after cross-hypothesis max-pooling operation.}
\label{fig:maxpooling}
\end{figure}

Based on the trained HCP model, the multi-label classification of a given image can be summarized as follows. We firstly generate the input hypotheses of the given image based on the proposed HS method. Then, for each hypothesis, a $c$-dimensional predictive result can be obtained by the shared CNN. Finally, we utilize the cross-hypothesis max-pooling operation to produce the final prediction. As shown in Fig.~\ref{fig:maxpooling}, the second row and the third row indicate the generated hypotheses and the corresponding outputs from the shared CNN. For each object independent hypothesis, there is a high response on the corresponding category (e.g., for the first hypothesis, the response on \emph{car} is very high). After cross-hypothesis max-pooling operation, as indicated by the last row in Fig.~\ref{fig:maxpooling}, the high responses (i.e., \emph{car}, \emph{horse} and \emph{person}), which can be considered as the predicted labels, are reserved.

\section{Experimental Results}
\label{sec:experiments}
In this section, we present the experiments to validate the effectiveness of our proposed Hypotheses-CNN-Pooling (HCP) framework for multi-label image classification.
\subsection{Datasets and Settings}
We evaluate the proposed HCP on the PASCAL Visual Object Classes Challenge (VOC) datasets~\cite{2010-pascal}, which are widely used as the benchmark for multi-label classification. In this paper, PASCAL VOC 2007 and VOC 2012 are employed for experiments. These two datasets, which contain 9,963 and 22,531 images respectively, are divided into \emph{train}, \emph{val} and \emph{test} subsets. We conduct our experiments on the \emph{trainval}/\emph{test} splits (5,011/4,952 for VOC 2007 and 11,540/10,991 for VOC 2012). The evaluation metric is \emph{Average Precision} (AP) and mean of AP (mAP) complying with the PASCAL challenge protocols. 


Instead of using two GPUs as in~\cite{2012-krizhevsky}, we conduct experiments on one NVIDIA GTX Titan GPU with 6GB memory and all our training algorithms are based on the code provided by Jia~\emph{et al.}~\cite{2013-Yangqing-caffe}. The initialization of the shared CNN is based on the parameters pre-trained on the 1,000 classes and 1.2 million images of ILSVRC-2012. 

We compare the proposed HCP with the state-of-the-art approaches. Specifically, the competing algorithms are generally divided into two types: those based on hand-crafted features and those based on \emph{learnt} features.
\begin{itemize}
\item[$\bullet$] \textbf{INRIA~\cite{2009-harzallah}:} Harzallah \emph{et al.} proposed a contextual combination method of localization and classification to improve the performance for both. Specifically, for classification, image representation is built on the traditional feature extraction-coding-pooling pipeline, and object localization is built on sliding-widow approaches. Furthermore, the localization is employed to enhance the classification performance. \\
\item[$\bullet$] \textbf{FV~\cite{2010-perronnin-improving}:} The Fisher Vector representation of images can be considered as an extension of the bag-of-words. Some well-motivated strategies, e.g., L2 normalization, power normalization and spatial pyramids, are adopted over the original Fisher Vector to boost the classification accuracy. \\
\item[$\bullet$] \textbf{NUS:} In~\cite{2014-chen}, Chen \emph{et al.} presented an Ambiguity-guided Mixture Model (AMM) to seamlessly integrate external context features and object features for general classification, and then the contextualized SVM was further utilized to iteratively and mutually boost the performance of object classification and detection tasks. Dong \emph{et al.}~\cite{2013-dong-subcategory} proposed an Ambiguity Guided Subcategory (AGS) mining approach, which can be seamlessly integrated into an effective subcategory-aware object classification framework, to improve both detection and classification performance. The combined version of the above two, NUS-PSL~\cite{2012-Shuicheng} received the winner prizes of the classification task in PASCAL VOC 2010-2012. \\
\item[$\bullet$] \textbf{CNN-SVM~\cite{2014-cnnfeatures}:} OverFeat~\cite{2013-overfeat}, which obtained very competitive performance in the image classification task of ILSVRC 2013, was released by Sermanet~\emph{et al.} as a feature extractor. Razavian \emph{et al.}~\cite{2014-cnnfeatures} employed OverFeat, which is pre-trained on ImageNet, to get CNN activations as the off-the-shelf features. The state-of-the-art classification result on PASCAL VOC 2007 was achieved by using linear SVM classifiers over the 4,096 dimensional feature representation extracted from the 22$^{nd}$ layer of OverFeat.  \\
\item[$\bullet$] \textbf{I-FT:} The structure of the shared CNN follows that of Krizhevsky \emph{et al.}~\cite{2012-krizhevsky}. The shared CNN was first pre-trained on ImageNet, and then the last fully-connected layer was modified into 4096$\times$20, and the shared CNN was  re-trained with squared loss function on PASCAL VOC for multi-label classification.\\
\item[$\bullet$] \textbf{PRE-1000C and PRE-1512~\cite{2013-transferCNN}:} Oquab~\emph{et al.} proposed to transfer image representations learned with CNN on ImageNet to other visual recognition tasks with limited training data. The network has exactly the same architecture as in~\cite{2012-krizhevsky}. Firstly, the network is pre-trained on ImageNet. Then the first seven layers of CNN are fixed with the pre-trained parameters and the last fully-connected layer is replaced by two adaptation layers. Finally, the adaptation layers are trained with images from the target PASCAL VOC dataset. PRE-1000C and PRE-1512 mean the transferred parameters are pre-trained on the original ImageNet dataset with 1000 categories and the augmented one with 1512 categories, respectively. For PRE-1000C, 1.2 million images from ILSVRC-2012 are employed to pre-train the CNN, while for PRE-1512, 512 additional ImageNet classes (e.g., \emph{furniture, motor vehicle, bicycle etc.}) are augmented to increase the semantic overlap with categories in PASCAL VOC. To accommodate the larger number of classes, the dimensions of the first two fully-connected layers are increased from 4,096 to 6,144.
\end{itemize}
\begin{figure}[t]
\centering
\includegraphics[scale=0.48]{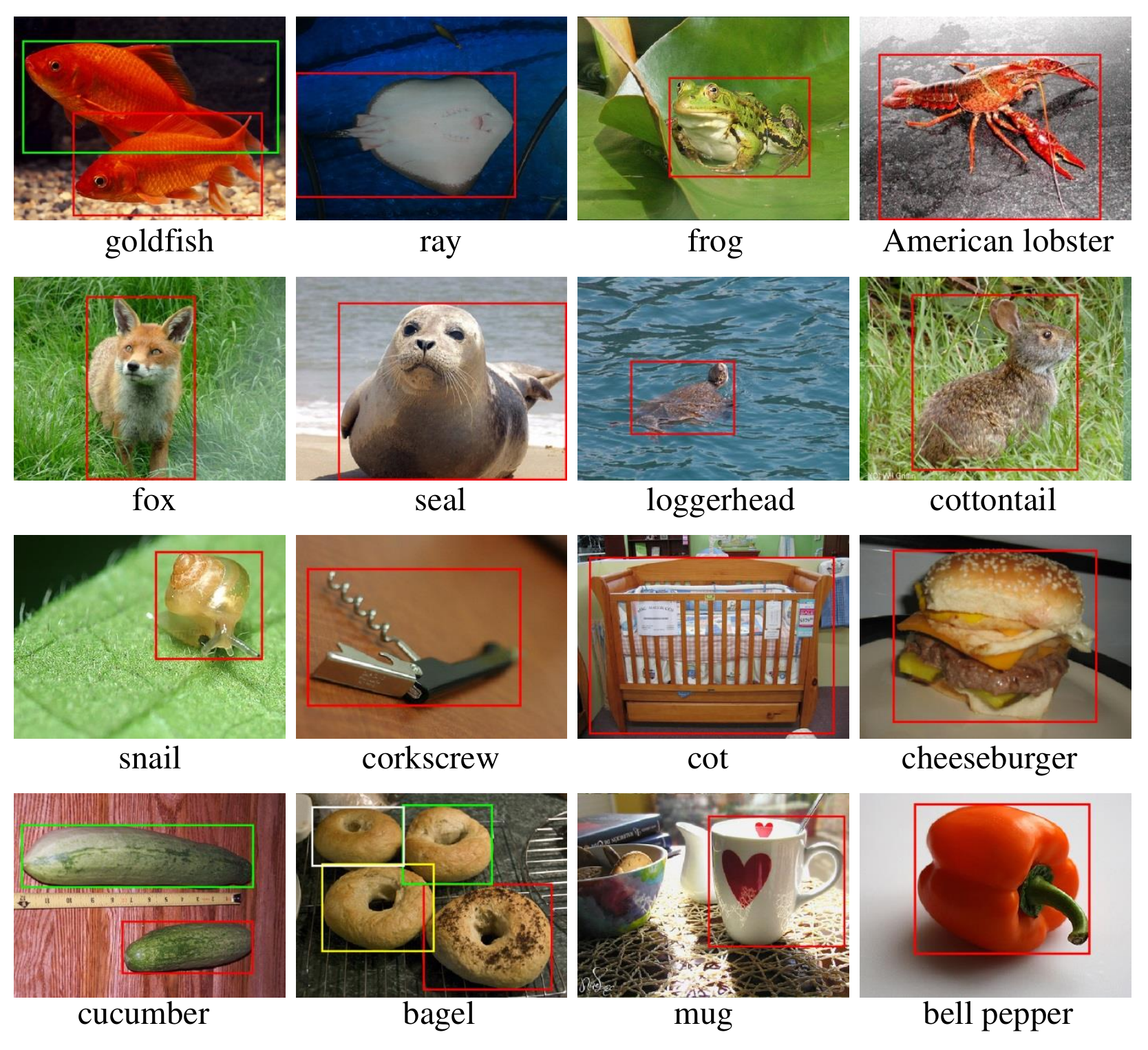}
\caption{Exemplar images with ground-truth bounding boxes from the detection dataset of ILSVRC 2013.}
\label{fig:imagenetclass}
\end{figure}

\subsection{Hypotheses Extraction}
In~\cite{2014-chengmm}, BING\footnote{http://mmcheng.net/bing/} has shown a good generalization ability on the images containing object categories that are not used for training. Specifically,~\cite{2014-chengmm} trained a linear SVM using 6 object categories (i.e., the first 6 categories in VOC dataset according to alpha order) and the remaining 14 categories were used for testing. The experimental results in~\cite{2014-chengmm} demonstrate that the transferred model almost has the same performance with that using all categories (all the 20 categories in PASCAL VOC) for training.

Since the proposed HCP is independent of the ground-truth bounding box, no object location information can be used for training. Inspired by the generalization ability test in~\cite{2014-chengmm}, the detection dataset of ILSVRC 2013 is used as augmented data for BING training. It contains 395,909 training images with ground-truth bounding box annotation from 200 categories. To validate the generalization ability of the proposed framework for other multi-label datasets, the categories as well as their subcategories which are semantically overlapping with the PASCAL VOC categoires are removed\footnote{The removed categories include \emph{bicycle, bird, water bottle, bus, car, domestic cat, chair, table, dog, horse, motorcycle, person, flower pot, sheep, sofa, train, tv or monitor, wine bottle, watercraft, unicycle, cattle and cart.}}.

For a fair comparison, we follow~\cite{2014-chengmm} and only use randomly 13,894 images (instead of all images) from the detection dataset of ILSVRC 2013 for model training. Some selected samples are illustrated in Figure~\ref{fig:imagenetclass}, from which we can see that there are big differences between objects in PASCAL VOC and the selected ImageNet samples. After training, the learnt BING-ImageNet model is used to produce hypotheses for VOC 2007 and VOC 2012. We test the object detection rate with 1000 proposals on VOC 2007, which is only 0.3\% lower than the reported result (i.e., 96.2\%) in~\cite{2014-chengmm}.

\begin{figure*}
\centering
\includegraphics[scale=0.43]{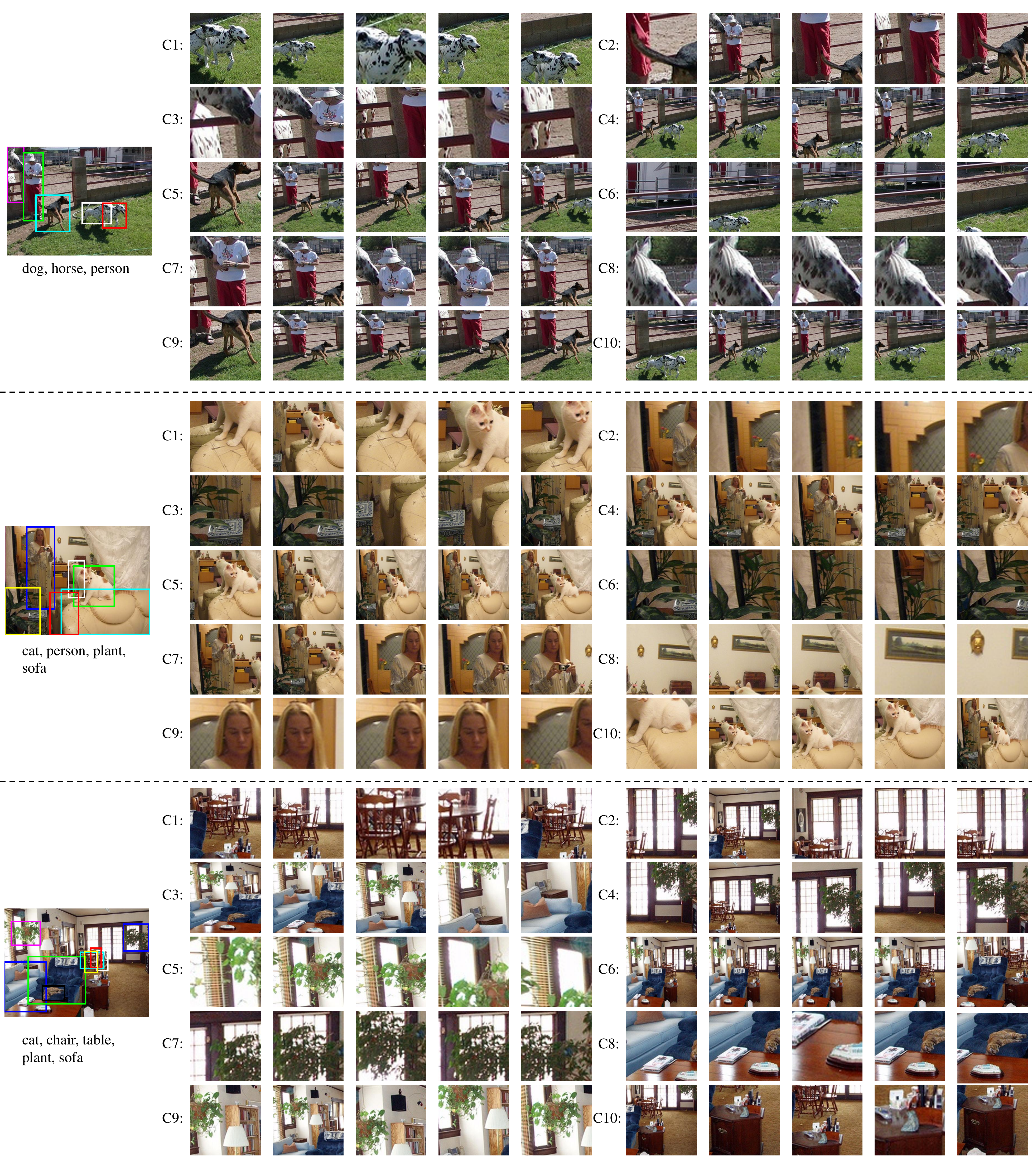}
\caption{Exemplar hypotheses extracted by the proposed HS method. For each image, the ground-truth bounding boxes are shown on the left and the corresponding hypotheses are shown on the right. C1-C10 are the 10 clusters produced by normalized-cut~\cite{2000-normalized} and hypotheses in the same cluster share similar location information. }
\label{fig:eg-proposals}
\end{figure*}

Considering the computational time and the limitation of hardware, we proposed a hypotheses selection (HS) method to filter the input hypotheses produced by BING-ImageNet. As elaborated in Section~\ref{subsec:hypotheses extraction}, we cluster the extracted proposals into 10 clusters based on their bounding box overlapping information by normalized-cut~\cite{2000-normalized}. The hypotheses are filtered out, which have smaller areas than 900 pixels or larger height/width (or width/height) ratios than 4. Some exemplar hypotheses extracted by the proposed HS method are shown in Figure~\ref{fig:eg-proposals}. We sort the hypotheses for each cluster based on the predicted objectness scores and show the first five hypotheses.

During the training step, for each training image, the top $k$ hypotheses from each cluster are selected and fed into the shared CNN. We experimentally vary $k = 1,2,3,4,5$ to train the proposed HCP on VOC 2007 and observe that the performance changes only slightly on the testing dataset. Therefore, we set $k=1$ (i.e., 10 hypotheses for each image) for both VOC 2007 and VOC 2012 during the training stage to reduce the training time. To achieve high object recall rate, 500 hypotheses (i.e., $k=50$) are extracted from each test image during the testing stage. On VOC 2012, the hypotheses-fine-tuning step takes roughly 20 hours. For each testing image, about 1.5 second is cost.

\begin{figure}[t]
\centering
\includegraphics[scale=0.48]{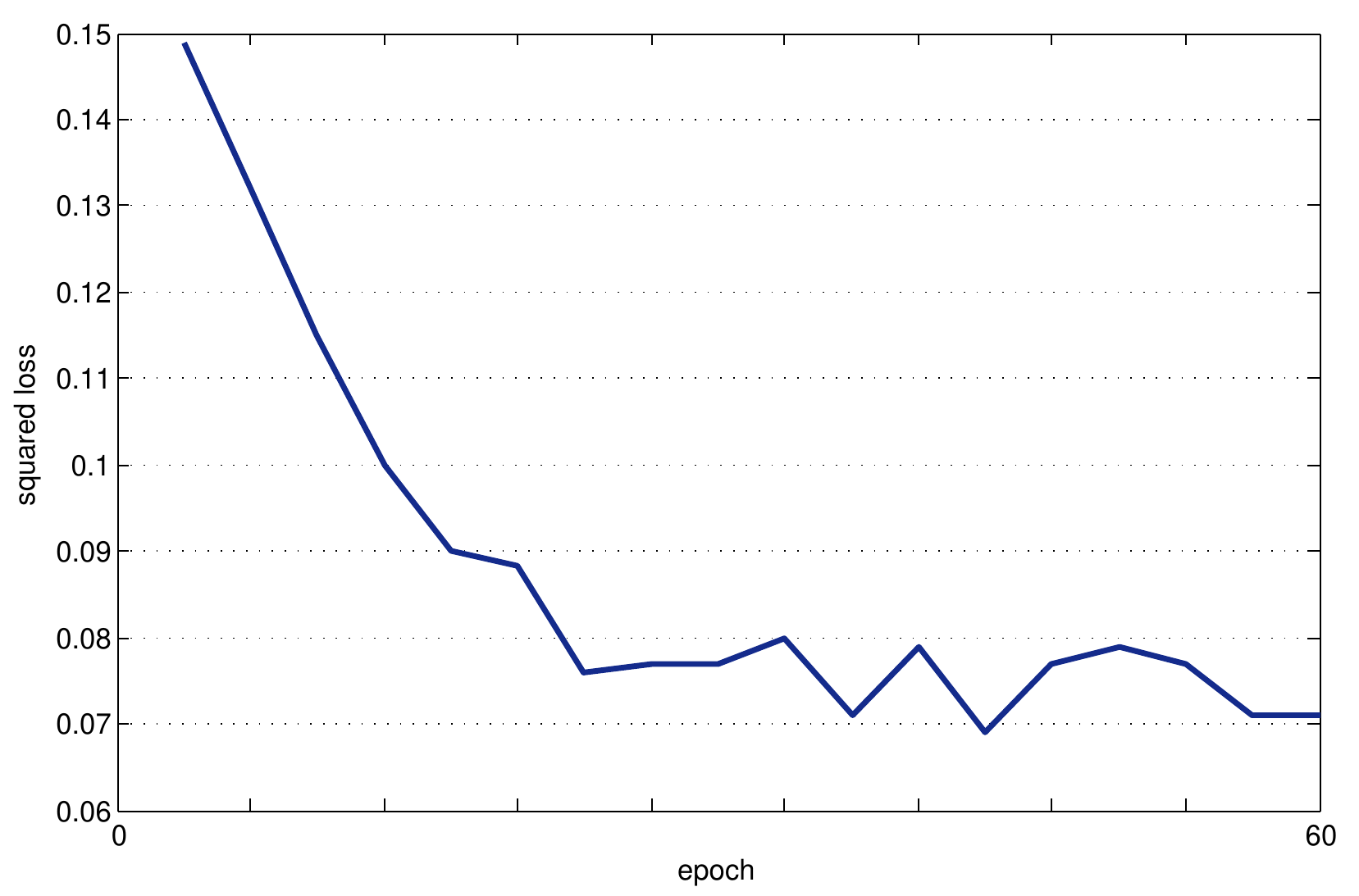}
\caption{The changing trend of the squared loss during the I-FT step on VOC 2007.  }
\label{fig:finetune_loss}
\end{figure}

\begin{figure}[t]
\centering
\includegraphics[scale=0.485]{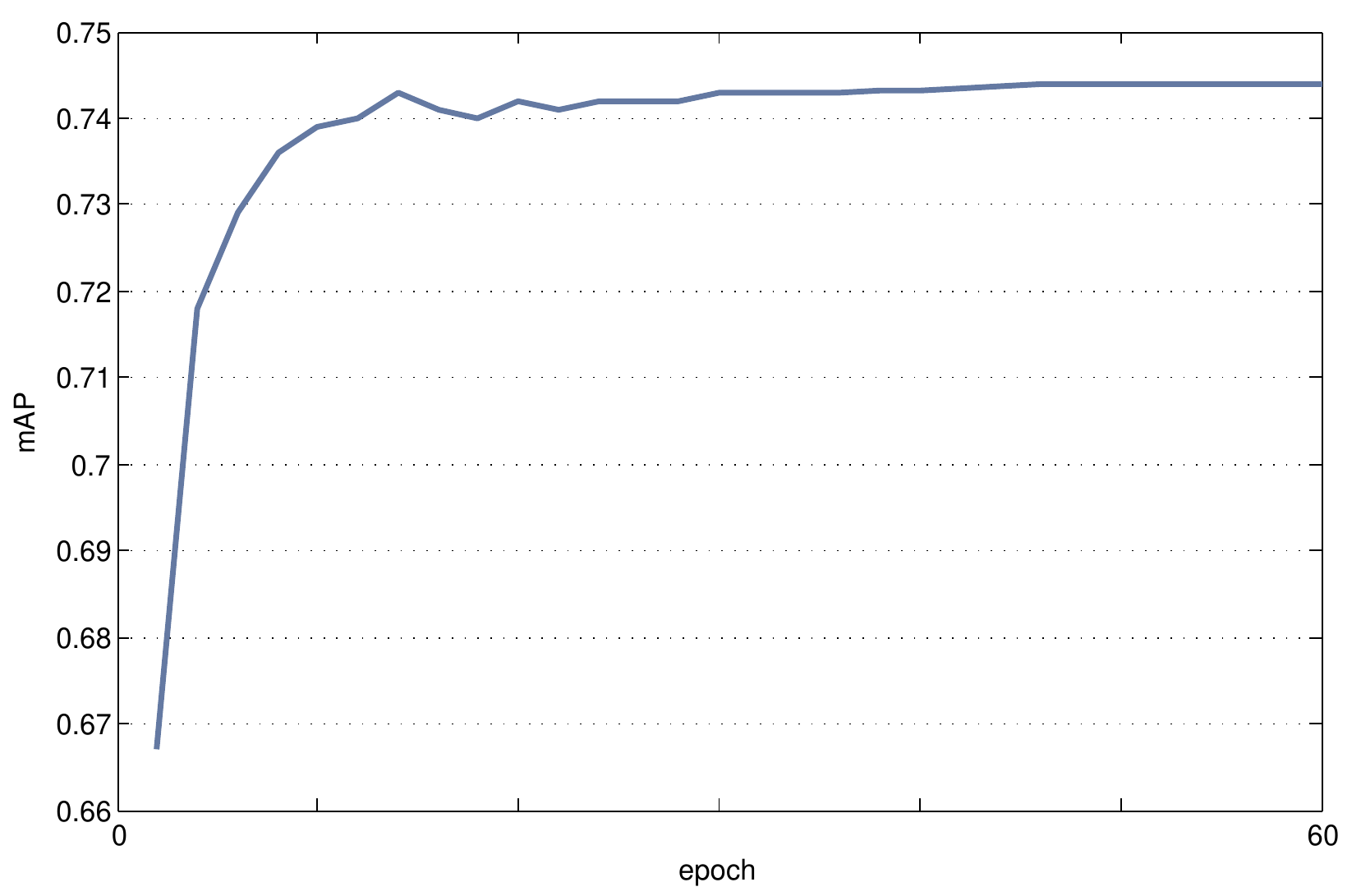}
\caption{The changing trend of mAP scores during the I-FT step on VOC 2007. The mAP converges fast to 74.4\% after almost 15 epoches on \emph{test} dataset. }
\label{fig:finetune_test}
\end{figure}


\subsection{Initialization of HCP}
As discussed in Section~\ref{subsection: initializeHCP}, the initialization process of HCP consists of two steps: Pre-training and Image-fine-tuning (I-FT). Since the structure setting of the shared-CNN is almost consistent with the pre-trained model implemented by~\cite{2013-Yangqing-caffe}, we apply the pre-trained model on ImageNet by~\cite{2013-Yangqing-caffe} to initialize the convolutional layers and the first two fully-connected layers of the shared CNN. For I-FT, we use images from PASCAL VOC to re-train the shared CNN. As shown in Figure~\ref{fig:finetune}, the pre-trained parameters for the first seven layers are transferred to initialize the CNN for fine-tuning. The last fully-connected layer with 4096$\times$20 parameters is randomly initialized with Gaussian distribution.

Actually, similar as Gong \emph{et al.}~\cite{2014-yunchao}, the class labels of a given image can also be predicted by the fine-tuned model at the I-FT stage. Figure~\ref{fig:finetune_loss} shows the changing trends of the squared loss of different epoches on VOC 2007 during I-FT. The corresponding change of mAP score on the \emph{test} dataset is shown in Figure~\ref{fig:finetune_test}. We can see that the mAP score based on the image-fine-tuned model can achieve 74.4\% on VOC 2007, which is more competitive than the scheme of CNN features with SVM classifier~\cite{2014-cnnfeatures}. 

\begin{table*}[htbp]\setlength{\tabcolsep}{1.2pt}
  \centering
  \caption{Classification results (AP in \%) comparison for state-of-the-art approaches on VOC 2007 (trainval/test). The upper part shows methods not using ground-truth bounding box information for training, while the lower part shows methods with that information. * indicates methods using additional data (i.e., ImageNet) for training.}
    \begin{tabular}{cccccccccccccccccccccc}
    \toprule
          & plane & bike  & bird  & boat  & bottle & bus   & car   & cat   & chair & cow   & table & dog   & horse & motor & person & plant & sheep & sofa  & train & tv    & mAP \\
    \midrule
    INRIA\cite{2009-harzallah} & 77.2  & 69.3  & 56.2  & 66.6  & 45.5  & 68.1  & 83.4  & 53.6  & 58.3  & 51.1  & 62.2  & 45.2  & 78.4  & 69.7  & 86.1  & 52.4  & 54.4  & 54.3  & 75.8  & 62.1  & 63.5 \\
    
    FV\cite{2010-perronnin-improving}    & 75.7  & 64.8  & 52.8  & 70.6  & 30.0  & 64.1  & 77.5  & 55.5  & 55.6  & 41.8  & 56.3  & 41.7  & 76.3  & 64.4  & 82.7  & 28.3  & 39.7  & 56.6  & 79.7  & 51.5  & 58.3 \\
    
    CNN-SVM*\cite{2014-cnnfeatures} & 88.5  & 81.0  & 83.5  & 82.0  & 42.0  & 72.5  & 85.3  & 81.6  & 59.9  & 58.5  & 66.5  & 77.8  & 81.8  & 78.8  & 90.2  & 54.8  & 71.1  & 62.6  & 87.4  & 71.8  & 73.9 \\
    
    I-FT* & 91.4  & 84.7  & 87.5  & 81.8  & 40.2  & 73.0  & 86.4  & 84.8  & 51.8  & 63.9  & 67.9  & 82.7  & 84. 0  & 76.9  & 90.4  & 51.5  & 79.9  & 54.1  & 89.5  & 65.8  & 74.4 \\
    
    HCP-1000C*& 95.1   & 90.1   & 92.8   & 89.9   & 51.5   & 80.0   & 91.7   & 91.6   & 57.7   & 77.8   & 70.9   & 89.3   & 89.3   & 85.2   & 93.0   & 64.0   & 85.7   & 62.7   & 94.4   & 78.3   & 81.5 \\
    HCP-2000C*& $\bm{96.0}$   & $\bm{92.1}$   & $\bm{93.7}$   & $\bm{93.4}$   & $\bm{58.7}$   & $\bm{84.0}$   & $\bm{93.4}$   & $\bm{92.0}$   & $\bm{62.8}$   & $\bm{89.1}$   & $\bm{76.3}$   & $\bm{91.4}$   & $\bm{95.0}$   & $\bm{87.8}$   & $\bm{93.1}$   & $\bm{69.9}$   & $\bm{90.3}$   & $\bm{68.0}$   & $\bm{96.8}$   & $\bm{80.6}$   & $\bm{85.2}$ \\
    \midrule
    \midrule
          & plane & bike  & bird  & boat  & bottle & bus   & car   & cat   & chair & cow   & table & dog   & horse & motor & person & plant & sheep & sofa  & train & tv    & mAP \\
    \midrule    
    AGS\cite{2013-dong-subcategory}   & 82.2  & 83.0  & 58.4  & 76.1  & 56.4  & 77.5  & 88.8  & 69.1  & 62.2  & 61.8  & 64.2  & 51.3  & 85.4  & 80.2  & 91.1  & 48.1  & 61.7  & 67.7  & 86.3  & 70.9  & 71.1 \\
    
    AMM\cite{2014-chen}   & 84.5  & 81.5  & 65.0  & 71.4  & 52.2  & 76.2  & 87.2  & 68.5  & 63.8  & 55.8  & 65.8  & 55.6  & 84.8  & 77.0  & 91.1  & 55.2  & 60.0  & 69.7  & 83.6  & 77.0  & 71.3 \\
    
    PRE-1000C*\cite{2013-transferCNN} & 88.5  & 81.5  & 87.9  & 82.0  & 47.5  & 75.5  & 90.1  & 87.2  & 61.6  & 75.7  & 67.3  & 85.5  & 83.5  & 80.0  & 95.6  & 60.8  & 76.8  & 58.0  & 90.4  & 77.9  & 77.7 \\
    \bottomrule
    \end{tabular}%
  \label{tab:voc2007}%
\end{table*}%

\begin{table*}[htbp]\setlength{\tabcolsep}{0.5pt}
  \centering
  \caption{Classification results (AP in \%) comparison for state-of-the-art approaches on VOC 2012 (trainval/test). The upper part shows methods not using ground-truth bounding box information for training, while the lower part shows methods with that information. * indicates methods using additional data (i.e., ImageNet) for training.}
    \begin{tabular}{cccccccccccccccccccccc}
    \toprule
          & plane & bike  & bird  & boat  & bottle & bus   & car   & cat   & chair & cow   & table & dog   & horse & motor & person & plant & sheep & sofa  & train & tv    & mAP \\
    \midrule
    I-FT* & 94.6   & 74.3   & 87.8   & 80.2   & 50.1   & 82.0   & 73.7   & 90.1   & 60.6   & 69.9   & 62.7   & 86.9   & 78.7   & 81.4  & 90.5   & 45.9   & 77.5   & 49.3   & 88.5   & 69.2   & 74.7 \\
    LeCun-ICML*\cite{2013-lecun-icml} & 96.0  & 77.1  & 88.4  & 85.5  & 55.8  & 85.8  & 78.6  & 91.2  & 65.0  & 74.4  & 67.7  & 87.8  & 86.0  & 85.1  & 90.9  & 52.2  & 83.6  & 61.1  & 91.8  & 76.1  & 79.0 \\
    HCP-1000C*   & 97.7   & 83.0   & 93.2   & 87.2   & 59.6   & 88.2   & 81.9   & 94.7   & 66.9   & 81.6   & 68.0   & 93.0   & 88.2   & 87.7   & 92.7   & 59.0   & 85.1   & 55.4   & 93.0   & 77.2   & 81.7 \\
    HCP-2000C*   & 97.5   & 84.3   & 93.0   & 89.4   & 62.5   & 90.2   & 84.6   & 94.8   & 69.7   & 90.2   & 74.1   & 93.4   & 93.7   & 88.8   & 93.3   & 59.7   & 90.3   & 61.8   & 94.4   & 78.0   & 84.2 \\
    \midrule
    \midrule
          & plane & bike  & bird  & boat  & bottle & bus   & car   & cat   & chair & cow   & table & dog   & horse & motor & person & plant & sheep & sofa  & train & tv    & mAP \\
    \midrule
    NUS-PSL\cite{2012-Shuicheng} & 97.3  & 84.2  & 80.8  & 85.3  & 60.8  & 89.9  & 86.8  & 89.3  & 75.4  & 77.8  & 75.1  & 83.0  & 87.5  & 90.1  & 95.0  & 57.8  & 79.2  & 73.4  & 94.5  & 80.7  & 82.2 \\
    PRE-1000C*\cite{2013-transferCNN} & 93.5  & 78.4  & 87.7  & 80.9  & 57.3  & 85.0  & 81.6  & 89.4  & 66.9  & 73.8  & 62.0  & 89.5  & 83.2  & 87.6  & 95.8  & 61.4  & 79.0  & 54.3  & 88.0  & 78.3  & 78.7 \\
    PRE-1512*\cite{2013-transferCNN} & 94.6  & 82.9  & 88.2  & 84.1  & 60.3  & 89.0  & 84.4  & 90.7  & 72.1  & 86.8  & 69.0  & 92.1  & 93.4  & 88.6  & $\bm{96.1}$  & 64.3  & 86.6  & 62.3  & 91.1  & 79.8  & 82.8 \\
    HCP-2000C+NUS-PSL* & $\bm{98.9}$   & $\bm{91.8}$   & $\bm{94.8}$   & $\bm{92.4}$   & $\bm{72.6}$   & $\bm{95.0}$   & $\bm{91.8}$   & $\bm{97.4}$   & $\bm{85.2}$   & $\bm{92.9}$   & $\bm{83.1}$   & $\bm{96.0}$   & $\bm{96.6}$   & $\bm{96.1}$   & 94.9   & $\bm{68.4}$   & $\bm{92.0}$   & $\bm{79.6}$   & $\bm{97.3}$   & $\bm{88.5}$   & $\bm{90.3}$ \\
    \bottomrule
    \end{tabular}%
  \label{tab:voc2012}%
\end{table*}%

\subsection{Image Classification Results}
\noindent\textbf{Image Classification on VOC 2007:} Table~\ref{tab:voc2007} reports our experimental results compared with the state-of-the-arts on VOC 2007. The upper part of Table~\ref{tab:voc2007} shows the methods not using ground-truth bounding box information for training, while the lower part of the table shows the methods with that information. Besides, CNN-SVM, I-FT, HCP-1000C, HCP-2000C and PRE-1000C are methods using additional images for training from an extra dataset, i.e., ImageNet, and the other methods only utilize PASCAL VOC data for training. In specific, HCP-1000C indicates that the initialized parameters of the shared CNN are pre-trained on the 1.2 million images from 1000 categories of ILSVRC-2012. Similar as~\cite{2013-transferCNN}, for HCP-2000C, we augment the ILSVRC-2012 training set with additional 1,000 ImageNet classes (about 0.8 million images) to improve the semantic overlap with classes in the Pascal VOC dataset.

From the experimental results, we can see that the CNN based methods which utilize additional images from ImageNet have a 2.6\%$\sim$13.9\% improvement compared with the state-of-the-art methods based on hand-crafted features, i.e., 71.3\%~\cite{2014-chen}. By utilizing the ground-truth bounding box information, a remarkable improvement can be achieved for both deep learning based methods (PRE-1000C \emph{vs.} CNN-SVM and I-FT) and hand-crafted feature based methods (AGS and AMM \emph{vs.} INRIA and FV). However, bounding box annotation is quite costly. Therefore, approaches requiring ground-truth bounding boxes cannot be transferred to the datasets without such annotation. From Table~\ref{tab:voc2007}, it can be seen that the proposed HCP has a significant improvement compared with the state-of-the-art performance even without bounding box annotation i.e., 81.5\% \textit{vs.} 77.7\% (HCP-1000C \textit{vs.} PRE-1000C~\cite{2013-transferCNN}). Compared with HCP-1000C, 3.7\% improvement can be achieved by HCP-2000C. Since the proposed HCP requires no bounding box annotation, the proposed method has a much stronger generalization ability to new multi-label datasets.

\begin{figure*}[htbp]
\centering
\includegraphics[scale=0.45]{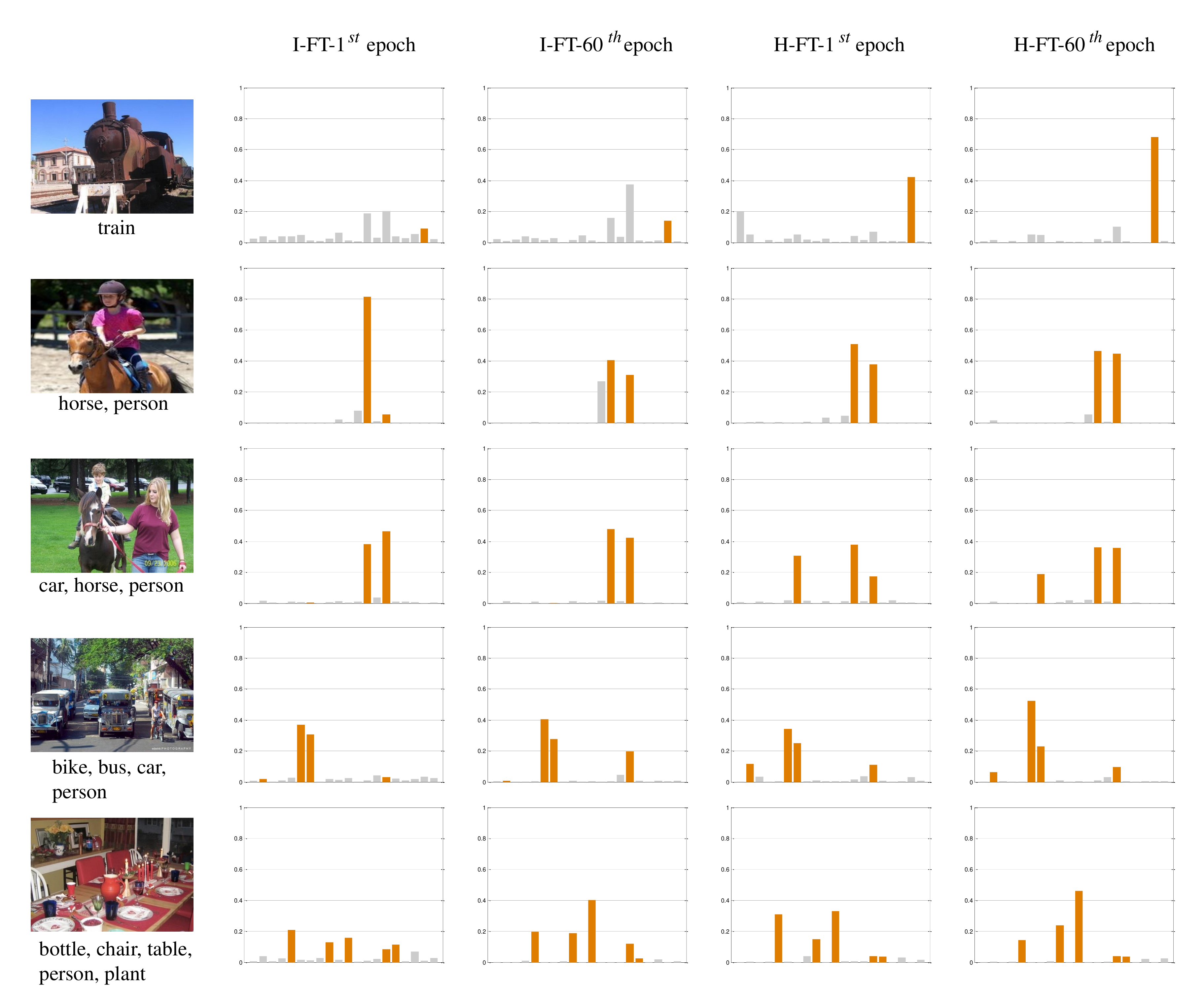}
\caption{Samples of predicted scores on the VOC 2007 testing dataset using models from different fine-tuning epochs (i.e., I-FT-$1^{st}$ epoch, I-FT-$60^{th}$ epoch, H-FT-$1^{st}$ epoch, and H-FT-$60^{th}$ epoch).}
\label{fig:featureshow}
\end{figure*}

Figure~\ref{fig:featureshow} shows the predicted scores of images for different categories on the VOC 2007 testing dataset using models from different fine-tuning epoches. For each histogram\footnote{For each histogram, categories from left to right are \emph{plane, bike, bird, boat, bottle, bus, car, cat, chair, cow, table, dog, horse, motor, person, plant, sheep, sofa, train and tv.}}, orange bars indicate predicted scores of the ground-truth categories. We show the predictive scores at the $1^{st}$ and the $60^{th}$ epoch during I-FT and H-FT stages. For the first row, it can be seen that the predictive score for the \emph{train} category gradually increases. Besides, for the third row, it can be seen that there are three ground-truth categories in the given image, i.e., \emph{car, horse, person}. It should be noted that the \emph{car} category is not detected during fine-tuning while it is successfully recovered in HCP. This may be because the proposed HCP is a hypotheses based method and both foreground (i.e., \emph{horse, person}) and background (i.e., \emph{car}) objects can be equivalently treated. However, during the fine-tuning stage, the entire image is treated as the input, which may lead to ignorance of some background categories.

\begin{table*}
\caption{Comparison in terms of rigid categories and articulated categories among NUS-PSL, PRE-1512 and HCP-2000C.}
\label{tab:comp}
\centering
\begin{tabular}{@{}ccccccccc@{}}
\toprule
\multirow{2}{*}{}            & \multirow{2}{*}{class} & \multicolumn{7}{c}{Comparison}                                                                                     \\ \cmidrule(l){3-9} 
                             &                        & NUS-PSL & PRE-1512 & HCP-2000C  & NUS-PSL \emph{vs.} PRE-1512 & mean                  & NUS-PSL \emph{vs.} HCP-2000C & mean                  \\ \midrule
\multirow{12}{*}{Rigid}      & plane                  & 97.3    & 94.6     & 97.5 & 2.7                 & \multirow{12}{*}{3.0} & -0.2           & \multirow{12}{*}{1.5} \\
                             & bike                   & 84.2    & 82.9     & 84.3 & 1.3                 &                        & -0.1            &                        \\
                             & boat                   & 85.3    & 84.1     & 89.4 & 1.2                 &                        & -4.1           &                        \\
                             & bottle                 & 60.8    & 60.3     & 62.5 & 0.5                 &                        & -1.7            &                        \\
                             & bus                    & 89.9    & 89.0     & 90.2 & 0.9                 &                        & -0.3            &                        \\
                             & car                    & 86.8    & 84.4     & 84.6 & 2.4                 &                        & 2.2            &                        \\
                             & chair                  & 75.4    & 72.1     & 69.7 & 3.3                 &                        & 5.7           &                        \\
                             & table                  & 75.1    & 69.0     & 74.1 & 6.1                 &                        & 1.0            &                        \\
                             & motor                  & 90.1    & 88.6     & 88.8 & 1.5                 &                        & 1.3            &                        \\
                             & sofa                   & 73.4    & 62.3     & 61.8 & 11.1                &                        & 11.6           &                        \\
                             & train                  & 94.5    & 91.1     & 94.4 & 3.4                 &                        & 0.1            &                        \\
                             & tv                     & 80.7    & 79.8     & 78.0 & 0.9                 &                        & 2.7            &                        \\ \midrule
\multirow{7}{*}{Articulated} & bird                   & 80.8    & 88.2     & 93.0 & -7.4                & \multirow{7}{*}{-5.9} & -12.2          & \multirow{7}{*}{-7.1} \\
                             & cat                    & 89.3    & 90.7     & 94.8 & -1.4                &                        & -5.5           &                        \\
                             & cow                    & 77.8    & 86.8     & 90.2 & -9.0                &                        & -12.4           &                        \\
                             & dog                    & 83.0    & 92.1     & 93.4 & -9.1                &                        & -10.4           &                        \\
                             & horse                  & 87.5    & 93.4     & 93.7 & -5.9                &                        & -6.2            &                        \\
                             & person                 & 95.0    & 96.1     & 93.3 & -1.1                &                        & 1.7            &                        \\
                             & sheep                  & 79.2    & 86.6     & 90.3 & -7.4                &                        & -11.1           &                        \\
                             & plant                  & 57.8    & 64.3     & 59.7 & -7.4                &                        & -0.9           &                        \\ \bottomrule
\end{tabular}
\end{table*}
\noindent\textbf{Image Classification on VOC 2012:} Table~\ref{tab:voc2012} reports our experimental results compared with the state-of-the-arts on VOC 2012. LeCun \emph{et al.}~\cite{2013-lecun-icml} reported the classification results on VOC 2012, which achieved the state-of-the-art performance without using any bounding box annotation. Compared with~\cite{2013-lecun-icml}, the proposed HCP-1000C has an improvement of 2.7\%. Both pre-trained on the ImageNet dataset with 1,000 classes, HCP-1000C gives a more competitive result compared with PRE-1000C~\cite{2013-transferCNN} (81.7\% \emph{vs.} 78.7\%). 

From Table~\ref{tab:voc2012}, it can be seen that the proposed HCP-1000C is not as competitive as NUS-PSL~\cite{2012-Shuicheng} and PRE-1512~\cite{2013-transferCNN}. This can be explained as follows. For NUS-PSL, which got the winner prize of the classification task in PASCAL VOC 2012, model fusion from both detection and classification is employed to generate the integrative result, while the proposed HCP-1000C is based on a single model without any fusion. For PRE-1512, 512 extra ImageNet classes are selected for CNN pre-training. In addition, the selected classes have intensive semantic overlap with PASCAL VOC, including \emph{hoofed mammal, furniture, motor vehicle, public transport, bicycle}. Therefore, the greater improvement of PRE-1512 compared with HCP-1000C is reasonable. By augmenting another 1,000 classes, our proposed HCP-2000C can achieve an improvement of 1.4\% compared with PRE-1512.

Finally, the comparison in terms of rigid and articulated categories among NUS-PSL, PRE-1512 and HCP-2000C is shown in Table~\ref{tab:comp}, from which it can be seen that the hand-crafted feature based scheme, i.e., NUS-PSL, outperforms almost all CNN feature based schemes for rigid categories, including \emph{plane, bike, boat, bottle, bus, car, chair, table, motor, sofa, train, tv}, while for articulated categories, CNN feature based schemes seem to be more powerful. Based on theses results, it can be observed that there is strong complementarity between hand-crafted feature based schemes and CNN feature based schemes. To verify this assumption, a late fusion between the predicted scores of NUS-PSL (also from the authors of this paper) and HCP is executed to make an enhanced prediction for VOC 2012. Incredibly, the mAP score on VOC 2012 can surge to 90.3\% as shown in Table~\ref{tab:voc2012}, which demonstrates the great complementarity between the traditional framework and the deep networks.

\section{Conclusions}
\label{sec:conclusions}
In this paper, we presented a novel Hypotheses-CNN-Pooling (HCP) framework to address the multi-label image classification problem. Based on the proposed HCP, CNN pre-trained on large-scale single-label image datasets, e.g., ImageNet, can be successfully transferred to tackle the multi-label problem. In addition, the proposed HCP requires no bounding box annotation for training, and thus can easily adapt to new multi-label datasets. We evaluated our method on VOC 2007 and VOC 2012, and verified that significant improvement can be made by HCP compared with the state-of-the-arts. Furthermore, it is proved that late fusion between outputs of CNN and hand-crafted feature schemes can incredibly enhance the classification performance. 
\ifCLASSOPTIONcaptionsoff
  \newpage
\fi

\bibliographystyle{ieee}
\bibliography{sigproc}
\end{document}